\documentclass[letterpaper]{article} % DO NOT CHANGE THIS
\usepackage{aaai23}  % DO NOT CHANGE THIS
\usepackage{times}  % DO NOT CHANGE THIS
\usepackage{helvet}  % DO NOT CHANGE THIS
\usepackage{courier}  % DO NOT CHANGE THIS
\usepackage[hyphens]{url}  % DO NOT CHANGE THIS
\usepackage{graphicx} % DO NOT CHANGE THIS
\usepackage{natbib}  % DO NOT CHANGE THIS AND DO NOT ADD ANY OPTIONS TO IT
\usepackage{caption} % DO NOT CHANGE THIS AND DO NOT ADD ANY OPTIONS TO IT
\usepackage{algorithm}
\usepackage{algorithmic}
\usepackage{graphicx}
\usepackage{tabularx}

\usepackage{newfloat}
\usepackage{listings}
\DeclareCaptionStyle{ruled}{labelfont=normalfont,labelsep=colon,strut=off} % DO NOT CHANGE THIS
\floatstyle{ruled}
\newfloat{listing}{tb}{lst}{}
\floatname{listing}{Listing}

\title{\centering\bfseries\fontsize{14}{16.8}\selectfont Leveraging AI-derived Data for Carbon Accounting: Information Extraction from Alternative Sources}

\author{
    Olamide Oladeji\textsuperscript{\rm 1,2}, Seyed Shahabeddin Mousavi\textsuperscript{\rm 1,2}\\\
}
\affiliations{
    \textsuperscript{\rm 1}Department of Management Science \& Engineering, Stanford University, Stanford, CA, USA\\
    \textsuperscript{\rm 2}Sustainable Finance Initiative, Precourt Institute for Energy, Stanford University, Stanford, CA, USA\\
    oladeji@stanford.edu, ssmousav@cs.stanford.edu
}

\usepackage{bibentry}
\begin{document}

\maketitle
\begin{abstract}
\begin{quote}
Carbon accounting is a fundamental building block in our global path to emissions reduction and decarbonization, yet many challenges exist in achieving reliable and trusted carbon accounting measures. We motivate that carbon accounting not only needs to be more data-driven, but also more methodologically sound. We discuss the need for alternative, more diverse data sources that can play a significant role on our path to trusted carbon accounting procedures and elaborate on not only why, but how Artificial Intelligence (AI) in general and Natural Language Processing (NLP) in particular can unlock reasonable access to a treasure trove of alternative data sets in light of the recent advances in the field that better enable the utilization of unstructured data in this process. We present a case study of the recent developments on real-world data via an NLP-powered analysis using OpenAI's GPT API on financial and shipping data. We conclude the paper with a discussion on how these methods and approaches can be integrated into a broader framework for AI-enabled integrative carbon accounting. 

\end{quote}
\end{abstract}

\noindent 

\section{Introduction}

The urgent need to mitigate climate change necessitates innovative approaches to track and manage greenhouse gas (GHG) emissions across various sectors, with effective carbon accounting being a critical step not only to track emissions but also for the achievement of global GHG emissions reduction goals. However, given the complex and dynamic nature of carbon emissions and sequestration, conventional methods often fall short in providing accurate, timely, and granular data \cite{brander2018creative} \cite{hayek2021underestimates}. The integration of Artificial Intelligence (AI) techniques can address these challenges, enhancing the efficacy and precision of carbon accounting.

AI-derived data, as gleaned through the application of advanced machine learning methods, offers us unprecedented opportunities for refined, dynamic, and real-time carbon accounting. In particular, Natural Language Processing (NLP) and Computer Vision (CV) stand out as promising AI techniques that can significantly enhance the world of carbon accounting. NLP techniques can aid in extracting meaningful data from voluminous and unstructured text, such as corporate filings or shipping records, that are imperative for understanding emissions across supply chains. At the same time, CV methods can enable the detection and tracking of significant carbon sources and sinks, such as wildfires and forests, via satellite imagery and remote sensing. Both techniques, especially when integrated into carbon accounting practices, can pave the way for more accurate and comprehensive monitoring and management of carbon emissions and sequestration across varied sectors and scales. This application of AI to carbon accounting is what we explore in this paper, with a particular emphasis on the illustrative use case of NLP for supply chain emissions tracking.

The work presented in this paper aims to demonstrate the efficacy of NLP in extracting emissions related information from supply chain textual sources, thereby facilitating the creation and continuous updating of a comprehensive knowledge graph for emissions tracking and management. Conventional methods such as manual documentation and reporting may sometimes be inadequate for this task given its scale and complexity as such mapping may need to be undertaken for an entire industry. It is the scalability that makes NLP entity recognition and relationship extraction algorithms a promising approach. The approach we present in this paper involves applying these techniques to earnings calls transcript data and shipping data with the goal of generating a dynamic, detailed model of supply chain e-liabilities and inform mitigation efforts. These approaches, when integrated with an appropriate E-liability knowledge graph framework, could transform our understanding and management of carbon emissions within supply chains, thereby facilitating effective emissions liability management. 

\subsection{Motivation and Related Work}
The complex task of managing carbon emissions within supply chains calls for a transformative approach that can process and extract insights from vast datasets. In this context, the utility of AI and Natural Language Processing (NLP) has been increasingly recognized \cite{stede2021climate} \cite{islam2022knowurenvironment}. NLP, in particular, with its capability to parse textual data and extract meaningful content, has been adopted across various domains, such as sentiment analysis \cite{loureiro2020sensing} and automated summarization \cite{islam2022knowurenvironment} \cite{gil2023using}.

With the advent of advanced language models such as OpenAI's GPT-4, the potential of NLP has become unprecedented, with recent models demonstrated comparable-to-human level competency on a wide variety of tasks across various fields \cite{sanderson2023gpt} \cite{bubeck2023sparks}. Supply chain management is one such area where the incorporation of NLP presents significant opportunities \cite{pournader2021artificial} \cite{schoenherr2015data}. Traditionally, supply chains have been characterized by a lack of transparency and a high degree of complexity\cite{ahi2013comparative}  and this opaqueness can pose significant challenges to effective carbon emissions tracking and management within the supply chain, a critical requirement in our fight against climate change \cite{sodhi2019research}. 

The literature has begun to explore the potential of NLP in enhancing supply chain transparency. For instance, \cite{schoenherr2015data} underscored the utility of NLP in risk assessment within the supply chain context. Jia et al. (2021) demonstrated how NLP can trace and monitor products throughout the supply chain, thereby fostering sustainability. Similarly, studies like \cite{zhou2021main} and \cite{wichmann2018towards} have also recognized the power of NLP in extracting valuable supply chain data. These solutions can best be used in conjunction with what Kaplan and Ramanna propose as the E-liability accounting system in \cite{kaplan2021accounting}, where they consider this system as one where emissions are measured using a combination of chemistry and engineering, and the principles of cost accounting are applied to assign the emissions to individual outputs. The authors provide a detailed method for assigning E-liabilities across an entire value chain, using the example of a car-door manufacturer whose furthest-removed supplier is a mining company, which transfers its products to a shipping company, which transports them to a steel company, and so on until the car reaches the end customer.

\section{Dataset}
We use two sources of data for this analysis: earnings calls and US shipping data. We discuss these sources as follows:

\begin{itemize}
\item \textbf{Earnings Calls:} The first data source utilized are earnings calls transcripts of publicly-traded companies in the United States. Earnings calls are teleconferences held by these publicly-traded companies for the purpose of reporting their financial results for a specific reporting period (quarterly or annually). During these calls, executives of these companies present summaries of the company's operations, while also discussing various future strategy and fielding questions from analysts and investors. The calls can serve as massive sources of information about the company's business activities and for this reason, tend to be followed closely by investors, analysts, and others interested in the company's performance and prospects.
In the context of our work, transcripts of earnings calls can provide valuable information about a company's operations, supply chain, and any environmental and sustainability efforts or issues they may be facing. Transcripts of these calls are freely available by providers such as The Street, MarketInsider, and SeekingAlpha. Stanford University, through its Graduate School of Business, also provides access to these transcripts as sourced from CapitalIQ to its researchers. For this analysis, we obtained sample earnings calls transcripts for Apple, Ford, GE, IBM and Tesla, for 2021, and used them to test our proposed NLP entity extraction methodology. We utilized 6 pairs of input-output few-shot examples overall, plus 29 test data sentences extracted from  pre-processed dataset that contained 326 real earnings call transcripts from Apple, Ford, IBM, GE, and Tesla mixed with some LLM-generated sentences based on the extracted sentences to address limited sample issues. To obtain the pre-processed dataset, we had human annotation experts review the transcripts to identify relevant sentences with buyer-supply-item and combine them with the LLM-generated sentences. Temperature was set to 0.1 to reduce randomness/’creativity’ of the model since this is an NER task.

\item \textbf{Bill of Lading Records:} We also obtained a sample dataset on Bill of Lading Records, which are similarly publicly available records - subject to transparency laws like the US Freedom of Information Act -  that detail geographical and company information about shipments moving in and out of a country. Third-party services such as ImportYeti and Panjiva aggregate and clean this data from the government and then provide them for use in research or business strategy analyses. We obtained a sample of such a dataset covering Bill of Lading records for the month of December 2020. 
The sample Bill of Lading data obtained was already in structured tabular form, spanning columns such as shipper name, shipper address, consignee name, consignee address, arrival date, product name, quantity and weight. Given the already structured nature of this data set, computational work undertaken focused on a basic illustrative exercise showing potential insights that could be garnered from such data, especially when converted to graphical form. As an example, three rows of the dataset used can be seen in Table 1.

\end{itemize}

\begin{table*}[htbp]
    \centering
    \small
    \begin{tabularx}{\textwidth}{|X|X|X|X|X|}
        \hline
        Shipper Name & Consignee Name & Product Desc & Quantity & Weight \\
        \hline
        % Add your table data here
        % Example with long data in the Product Desc column:
        BAODING HUANGHENG BAGS MANUFACTURING CO LTD & YOUNG LIVING ESSENTIAL OILS & HANDBAG THIS SHIPMENT CONTAINS NO WOOD PACKAGING MATERIALS. & 20 & 990 \\
        \hline
        THANH CONG TEXTILE GARMENT INVESTMENT TRADE JSC & MIAS FASHION MANUFACTURING COMPANY INC & AERO WOMEN S SCOOP NECK BUNGEE CAMI AERO WOMEN S SHORT PANTS NAUTICA WOMEN NAUTICA HERITAGE DP APPLIQUE & 1445 & 2767 \\
        \hline
        PELTER WINERY LTD &  ISRAELI WINE DIRECT LLC & SLAC WINE ON 2 PACKAGES HS 220429 & 35 & 714 \\
        \hline
    \end{tabularx}
    \caption{Example Rows from the bill of ladings dataset}
\end{table*}

\section{Methodology}
Given the large volume of earnings call transcripts and the nature of the contents, we run an initial Named Entity Recognition (NER) model to filter out sentences which mention the name of a company. For this, we use spaCy, an open-source natural language processing library which has a number of pre-trained models and pipelines suitable for named entity recognition \cite{spacy}. In particular, the en\_core\_web\_sm model, an English model already trained on web text (blog, news, comments) was used. We use the spaCy NER model as a pre-processing step to pre-select paragraph samples related to different companies. This step allows us to filter out many sentence fillers and focus solely on sentences where another company might be mentioned or indirectly referred to. 
For the next step, we focus on using a state-of-the-art large language model (LLM) from OpenAI. Using the OpenAI GPT API, code was developed to undertake few shot learning on the davinci-003 model. To achieve this, we curated a list of paired input-output training examples for the model of text. Three samples of these paired input-output training examples for the model is as below:

\begin{itemize}
\item \textbf{Input \#1:} Amazon relies heavily on UPS for the delivery of its goods.
\item \textbf{Output \#1:} Buyer: Amazon, Supplier: UPS, Item: delivery services
\item \textbf{Input \#2:} Samsung's OLED screens are sourced mainly from LG Display.
\item \textbf{Output \#2:} Buyer: Samsung, Supplier: LG Display, Item: OLED screens
\item \textbf{Input \#3:} The leather for our shoes comes from a supplier in Italy.
\item \textbf{Output \#3:} Buyer: $<$Your company$>$, Supplier: $<$Italian supplier$>$, Item: leather

\end{itemize}

For the Bill of Lading data analysis, the limitations of the dataset means that we are unable to construct a full e-liability knowledge graph as this would require its combination with other data  such as comprehensive domestic purchases and process-emission data on companies. 

\begin{figure}[h!]
\centering
\includegraphics[width=\linewidth]{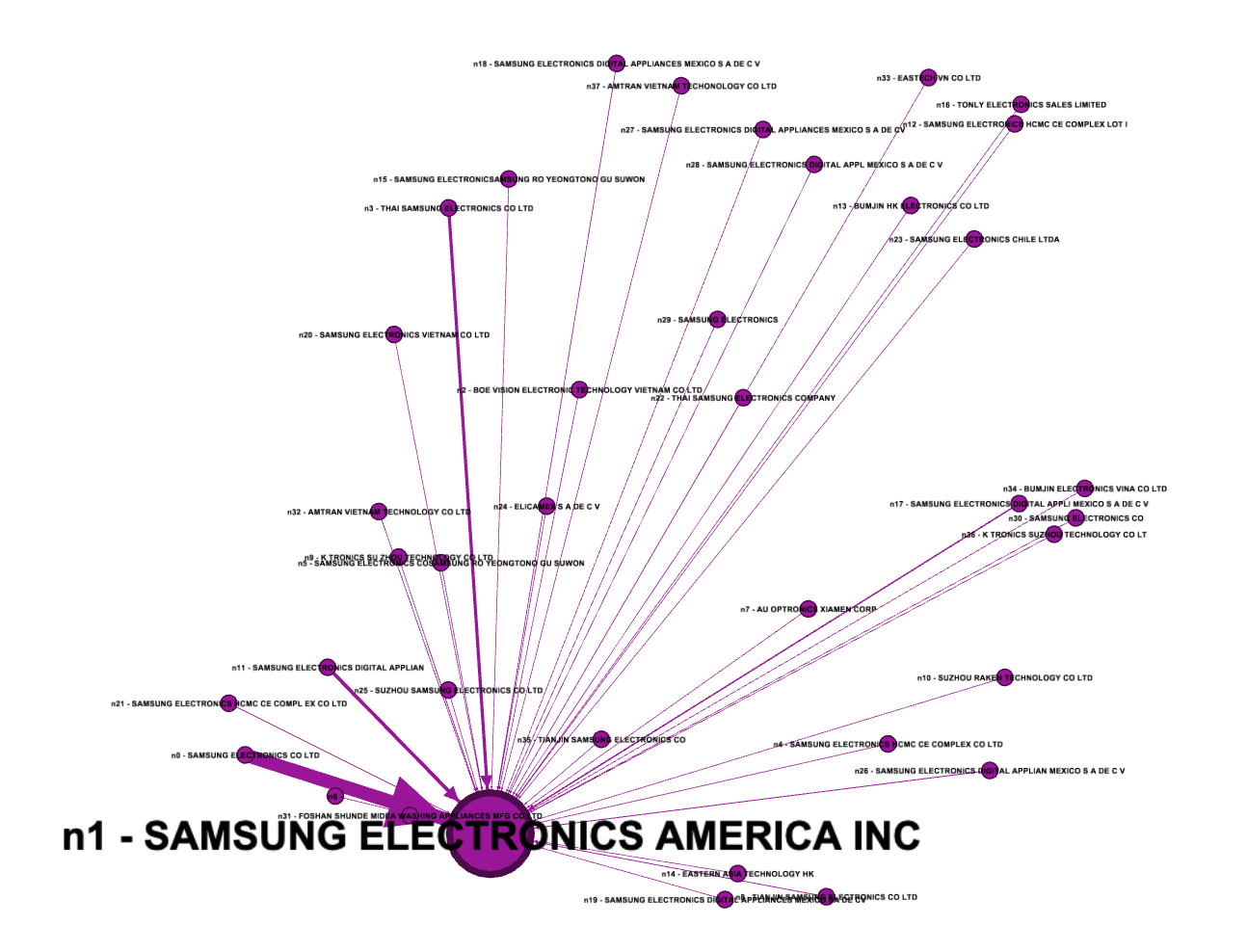}
\caption{Simplistic Graph showing E-liability inheritance for Samsung Electronics America Inc. using Bill of Lading data}
\label{samsung}
\end{figure}

Thus, our approach here was to subset the dataframe to specific popular companies and then identify shippers and items shipped. As an additional step, for illustrative visualization purposes, we generate product E-liabilities values for these shipments. We do this by sampling, from a Gaussian random variable distribution, the kg CO2-equivalent emitted per kg of each item shipped for these companies. We then multiply this sampled value with the weight data in the dataset. Instead of sampling, actual emissions for these products could have been estimated from Life Cycle Analysis (LCA) databases such as Eco-invent, however, accessing Eco-invents APIs proved to be prohibitively expensive (financially) considering our team's budget. For our purposes, we filter out a specific consignee, Samsung Electronics America Inc. and identify all shippers to this company mentioned in the Bill of Lading. In total, we identify 2261 shipments from 38 unique firms. Using iGraph and our random variable E-liabilities, we compute inherited E-liabilities for Samsung Electronics Americas due to these shipments and then visualize the overall flow in Gephi as shown in Figure \ref{samsung}. In Figure \ref{samsung}, the edges are weighted based on E-liabilities flow. 

\section{Results}

Using an 80:20 split with 80\% of the data used as example prompts for few shot learning and the rest used as the test set, we explain here the result of the model. Here, with rigorous training and fine-tuning, our GPT-3 (davinci-003) based Transformer-based Large Language Model (LLM) demonstrated an impressive performance in extracting key details related to supply chain transactions from pre-parsed company earnings call transcript sentences. The extraction task involved identifying the buyer, the supplier, and the item involved in each transaction.

Our model achieved a perfect score in terms of precision, recall, F1-score, and accuracy for buyer detection. It exhibited slight room for improvement in supplier and item detection, with a recall of approximately 0.958 and a corresponding F1-score of approximately 0.979, both indicative of the model's high precision and its capability to identify most actual instances of suppliers and items.

The performance of the model was optimized by adjusting the temperature parameter of the model's softmax function to 0.1. This temperature setting tends to make the model's outputs more deterministic and less diverse, which is beneficial for tasks like this that require exact matches and where the possible outputs are constrained.

The model's high performance demonstrates that state-of-the-art Natural Language Processing techniques - in particular - the Transformer-based LLMs like GPT, can be extremely effective in extracting complex relational information from text data. This capability  means that there is significant potential for the use of AI in developing knowledge graphs that track complex E-liabilities across supply chains, thus providing valuable insights for decision makers.

We did not run the model on full transcripts given the high volume of data and the computational cost of accessing OpenAI's APIs. However, the exemplary performance of the model on this subset of data show that this approach is very promising and amenable to high scalability if API costs were not an obstacle. 

We present below the results in tabular form in Table 1:

\begin{table}[h]
\centering
\begin{tabular}{|l|l|l|l|l|}
\hline
 & Precision & Recall & F1 Score & Accuracy \\ \hline
Buyer & 1.0 & 1.0 & 1.0 & 1.0 \\ \hline
Supplier & 1.0 & 0.958 & 0.979 & 0.958 \\ \hline
Item & 1.0 & 0.958 & 0.979 & 0.958 \\ \hline
\end{tabular}
\caption{Performance of GPT (davinci-003) on extracting buyers, suppliers, and items from text.}
\end{table}

\section{Conclusion}

In the carbon accounting literature, it is established that entities only need to track the E-liabilities associated with their own processes alongside the E-liabilities they inherit from their suppliers in order to estimate their total E-liabilities. \cite{von2022pulling}
We have also demonstrated that using alternative data sources, such as Bill of Lading information and earnings calls transcripts, we can garner, with reasonable accuracy, multiple insights into a company's supply chain dynamics, gaining a better understanding  of items purchased or supplied and other relevant transaction details. We can then combine such insights with product life cycle emissions inventories to construct alternative e-liability knowledge graphs that map the trajectory and accumulation of E-liabilities across industries and geographies. Such an alternative, indirect approach to building elements of e-liability knowledge graphs at the industry scale can be useful in several ways. We discuss a few of these below:

\begin{itemize}
    \item
    \textbf{Verification/Independent Auditing of Carbon Accounts:}
Public companies in particular generate lots of unstructured text data related to their operational activity, whether through SEC filings, earnings calls or Bill of Lading information. By mining such text to construct a knowledge graph on their E-liabilities, independent auditors can more easily detect any false claims in E-liabilities. 

\item 
\textbf{Industrial Benchmarking:} Such datasets and the approaches outlined previously can be combined and scaled to industry levels, allowing decision makers and researchers to easily generate industrial benchmarks and to compare the performance of companies across the board. 

\item 
\textbf{Research and Development:} By mining sources in the approach we have presented, researchers are better enabled to conduct research to identify both areas with the greatest potential for mitigation and pitfalls faced on our path to comprehensive and reliable carbon accounting. Industrial-scale e-liability knowledge graphs can facilitate the identification of high impact processes or industries.

\item 
\textbf{Policy-Making and Impact Analysis:} Using such datasets, policymakers can more easily formulate effective policies around carbon emissions. For example, they are better able to target subsectors with the highest E-liabilities as they do not necessarily need to wait to obtain silo-ed E-liabilities data. They can also measure the impact of several regulations by monitoring changes in the knowledge graph over time. The effect of a carbon price regulation on emissions across an industry can be easily tracked on the e-liability knowledge graph constructed from alternative sources. Similarly, using geographically constructed knowledge E-liabilities, policymakers can identify regions with high emissions and implement regional specific policies. 
 \end{itemize}

\bibliographystyle{aaai} \bibliography{aaai23.bib}

\end{document}